# Dimensionality Reduction by Local Discriminative Gaussians

**Nathan Parrish**                                                            NPARRISH@U.WASHINGTON.EDU

University of Washington, Department of Electrical Engineering, Seattle, WA 98195 USA

**Maya R. Gupta**                                                             GUPTA@EE.WASHINGTON.EDU

University of Washington, Department of Electrical Engineering, Seattle, WA 98195 USA

## Abstract

We present local discriminative Gaussian (LDG) dimensionality reduction, a supervised dimensionality reduction technique for classification. The LDG objective function is an approximation to the leave-one-out training error of a local quadratic discriminant analysis classifier, and thus acts locally to each training point in order to find a mapping where similar data can be discriminated from dissimilar data. While other state-of-the-art linear dimensionality reduction methods require gradient descent or iterative solution approaches, LDG is solved with a single eigen-decomposition. Thus, it scales better for datasets with a large number of feature dimensions or training examples. We also adapt LDG to the transfer learning setting, and show that it achieves good performance when the test data distribution differs from that of the training data.

## 1. Introduction

Dimensionality reduction is the mapping of high-dimensional data into a lower-dimensional space while retaining as much of the information content of the data as possible. As a preprocessing step for supervised classification algorithms, dimensionality reduction achieves several important goals. It reduces the storage requirements and algorithm complexity by reducing the input space of the data. It can improve performance of learning algorithms by rejecting spurious or noisy features prior to training and testing. Dimensionality reduction can also protect against overfitting by reducing the number of parameters learned by the classifier.

We present a method for supervised dimensionality reduction that is based on a local discriminative Gaussian (LDG) criterion. The discriminative Gaussian criterion is a smooth approximation to the leave-one-out cross-validation error of a quadratic discriminant analysis (QDA) classifier, so it seeks a mapping where a quadratic boundary separates the classes. Because this goal of separation by class may be difficult to achieve globally, our criterion instead operates locally to each training point.

The considered objective function is non-convex with no analytical solution; however, we present an approximation that is solved via a maximal eigenvalue decomposition. The simplicity of the solution is an advantage over other state-of-the-art dimensionality reduction techniques that require iterative solution methods or more complex generalized eigenvalue decompositions.

We perform experiments for supervised dimensionality reduction, and for dimensionality reduction for transfer learning. We show that on datasets with a large number of feature dimensions, other state-of-the-art algorithms are either intractably slow or exhibit numerical instability, whereas LDG is able to extract a useful mapping even when the number of features in the original data is in the thousands. We also show that LDG can be easily extended to the transfer learning setting, where the training data is drawn from a different distribution than the test data. Experiments show that LDG is effective in this setting as well.

## 2. Problem Formulation

We take as given a set of labeled training data $\{(x_i, y_i)\}_{i=1}^{n}$, with $x_i \in \mathbb{R}^d$ and $y_i \in \{1, 2, ..., m\}$ being the $i^{\text{th}}$ feature vector and class label respectively.

We wish to find a matrix $B \in \mathbb{R}^{d \times l}, l < d$ such that the reduced-dimensionality feature vectors $\{B^T x_i\}$ can be separated according to class. We measure this separability by the performance of a generative classifier. Let $p(x_i|y_i)$ be the likelihood of $x_i$ given class $y_i$, estimated from the other $n-1$ training sample pairs. Then the leave-one-out cross-validation error of a maximum a-posteriori (MAP) classifier acting on the mapped features measures the separation achieved by $B$:

$$\sum_{i=1}^n \mathrm{I}\left(p(B^T x_i|y_i)p(y_i) < \max_j p(B^T x_i|j)p(j)\right), \quad (1)$$

where the indicator function $\mathrm{I}(\cdot)$ is one if its argument is true and zero otherwise.

The discontinuity of the indicator function in (1) makes it difficult to minimize. In order to arrive at a smooth, differentiable objective function that approximates (1), we substitute a log for the indicator and a sum for the max:

$$f(B) = \sum_{i=1}^n \log\left(\frac{\sum_{j=1}^m p(B^T x_i|j)p(j)}{p(B^T x_i|y_i)p(y_i)}\right). \quad (2)$$

In related work, $p(x_i|j)$ was assumed to be a Gaussian mixture model (GMM), and the objective was to learn a parameter vector, $\Theta$, of GMM weights, means, and variances that minimized $f(\Theta)$ with $p(x_i|j,\Theta)$ replacing $p(B^T x_i|j)$ in (2) (Ma & Chang, 2003). The learned parameters were shown to improve the GMM classification performance over the parameters learned by maximimum likelihood estimation. In that work, (2) is motivated as maximizing the mutual information between the class labels and the feature vectors.

We assume $p(x_i|j)$ is Gaussian, $\mathcal{N}(x_i; \mu_{i,j}, \Sigma_{i,j})$. However, to reduce the model bias of assuming one Gaussian per class, we model $p(x_i|j)$ as *locally* Gaussian (Garcia et al., 2010). That is, we estimate the parameters of the Gaussian for point $x_i$ and class $j$ by finding the $k$ nearest class $j$ neighbors, in Euclidean distance, to training point $x_i$ and using these points to estimate the Gaussian's maximum likelihood mean and covariance. To reduce estimation variance, we model each covariance matrix as a scaled identity $\Sigma_{i,j} = \sigma_{i,j}^2 I$, where $I$ is the properly sized identity matrix. Therefore, $p(B^T x_i|j) = \mathcal{N}(B^T x_i; B^T \mu_{i,j}, B^T B \sigma_{i,j}^2)$.

Objective (2) is non-convex with no analytical solution. Gradient-descent or global optimization can be used, but become computationally expensive if the number of classes, training samples, or dimensionality are large. Therefore, we propose a tractable approximation that has an analytical solution. The $B$ that minimizes our approximation can be used directly (as we do in our experiments), or as a starting point for a gradient descent approach to minimizing (2).

We rewrite (2) as

$$f(B) = \sum_{i=1}^n \left(\log\left(\sum_{j=1}^m p(B^T x_i|j)p(j)\right) \quad (3)\right.$$
$$\left. - \log\left(p(B^T x_i|y_i)p(y_i)\right)\right),$$

and bound (2) from below with Jensen's inequality by replacing the first log term in (3) with $\sum_{j=1}^m p(j)\log(p(B^T x_i|j))$. Also, we impose the constraint that $B^T B = I$. This constraint simplifies (2) by making the covariance of the Gaussians in the mapped space independent of $B$. Furthermore, it makes for a unique solution. After taking the log of the Gaussians, we arrive at the LDG objective:

$$B^* = \arg\min_{B \in \mathbb{R}^{d \times l}} \sum_{i=1}^n \left(\frac{1 - p(y_i)}{2\sigma_{i,y_i}^2} \Delta_{i,y_i}^T BB^T \Delta_{i,y_i} \quad (4)\right.$$
$$\left. - \sum_{j=1, j \neq y_i}^m \left(\frac{p(j)}{2\sigma_{i,j}^2} \Delta_{i,j}^T BB^T \Delta_{i,j}\right)\right)$$
$$\text{s.t. } B^T B = I.$$

where $\Delta_{i,j} = \mu_{i,j} - x_i$.

Despite the approximations, (4) retains an intuitive meaning. The $B$ that minimizes the first term in (4) is the maximum likelihood solution for the *correct-class* local Gaussians. The second term is composed of $m-1$ different terms, each of which, if minimized individually, will give the maximum likelihood solution for an *incorrect-class* Gaussian, i.e. a Gaussian distribution trained by the local neighbors of $x_i$ coming from a different class. Therefore, (4) can be viewed as a regularized maximum likelihood estimate, where the regularization term attempts to minimize the likelihood of incorrect classes.

### 2.1. LDG Solution

The $B$ that optimizes (4) can be found with one eigendecomposition. Define

$$V = \sum_{i=1}^n \frac{1}{\sigma_{i,y_i}^2} \Delta_{i,y_i} \Delta_{i,y_i}^T$$
$$A = \sum_{i=1}^n \sum_{j=1}^m \frac{p(j)}{\sigma_{i,j}^2} \Delta_{i,j} \Delta_{i,j}^T.$$

Dimensionality Reduction by Local Discriminative Gaussians

Then (4) can be written as

$$B^* = \underset{B \in \mathbb{R}^{d \times l}}{\arg\min} \frac{1}{2} \text{Tr}\left(B^T(V-A)B\right) \quad (5)$$
$$\text{s.t. } B^T B = I.$$

Since both $V$ and $A$ are real symmetric matrices, it is straightforward to show that the solution to (5) is to set $B^*$'s columns to be the $l$ smallest eigenvectors of matrix $(V - A)$.

Additionally, we add a cross-validated regularization parameter, $\gamma$, to (4), which we have found in practice can produce a mapping that better separates the data:

$$B^*_\gamma = \underset{B \in \mathbb{R}^{d \times l}}{\arg\min} \sum_{i=1}^n \left( \frac{1}{2\sigma^2_{i,y_i}} \Delta^T_{i,y_i} BB^T \Delta_{i,y_i} \right. \quad (6)$$
$$\left. - \gamma \sum_{j=1}^m \left( \frac{p(j)}{2\sigma^2_{i,j}} \Delta^T_{i,j} BB^T \Delta_{i,j} \right) \right)$$
$$\text{s.t. } B^T B = I.$$

The solution to (6) is to set $B^*_\gamma$'s columns to be the $l$ smallest eigenvectors of matrix $(V - \gamma A)$.

## 3. Related Methods for Linear Dimensionality Reduction

Fisher discriminant analysis (FDA) (Fisher, 1936) is a supervised technique that chooses $B$ to maximize the ratio of the between-class covariance $S^{(b)}$ to the within-class covariance $S^{(w)}$. The solution is to choose the top eigenvectors of the generalized eigen-decomposition $S^{(b)} \lambda = \nu S^{(w)} \lambda$. FDA has two drawbacks. First, FDA can perform poorly on multi-modal data where no single linear boundary separates the data by class. Second, the between-class covariance matrix is at most rank $m - 1$, so FDA can provide at most $m - 1$ dimensions.

Local Fisher discriminant analysis (LFDA) (Sugiyama, 2007) alleviates the drawbacks of FDA. LFDA generalizes FDA by adding a weight based on pairwise sample distances to the between-class and within-class covariance matrices. Thus, LFDA is able to separate multi-modal data. This change also results in LFDA being able to provide greater than $m - 1$ dimensions. LFDA is solved using the same generalized eigen-decomposition as FDA.

Neighbourhood components analysis (NCA) (Globerson et al., 2005) is a dimensionality reduction technique that is based on a smooth approximation to the leave-one-out k-NN error. The dimensionality reduction found by NCA was shown to provide good classification accuracy; however, it suffers from two key drawbacks. First, the optimization requires gradient descent, and can be slow for datasets with a large number of features or training examples. Second, the NCA optimization must be re-run for any desired number of final dimensions. This is in contrast to principal components analysis (PCA), FDA, LFDA, and LDG, where $B$ can be found once for the largest number of final dimensions, and then the top submatrices of $B$ are the optimal solution for fewer dimensions.

Finally, there is a large body of work in distance metric learning and feature selection that is related to linear dimensionality reduction. Distance metric learning addresses the problem of how best to determine the distance between feature vectors in $\mathbb{R}^d$. Linear distance metric learning is primarily concerned with finding a positive semi-definite Mahalanobis metric $M$ that gives the distance between $x_i$ and $x_\ell$ as $\sqrt{(x_i - x_\ell)^T M (x_i - x_\ell)}$. Linear dimensionality reduction can be thought of as finding a low rank Mahalanobis metric $M$, such that $M = BB^T$. The approaches given in (Globerson & Roweis, 2006; Davis et al., 2007; Weinberger & Saul, 2009) propose convex optimization problems for finding $M$. These methods suffer from the drawback that rank constraints are non-convex, and thus the $M$ that they find is typically not low rank. However, we can perform dimensionality reduction by rewriting the Mahalanobis metric as $M = L\Lambda L^T$ and using a feature selection method on the resulting $z_i = \Lambda^{1/2} L^T x_i$ as proposed in (Globerson & Roweis, 2006; Davis et al., 2007).

## 4. Dimensionality Reduction Experiments

We perform experiments to compare LDG to several different dimensionality reduction methods: PCA, FDA, LFDA, NCA, and information theoretic metric learning (ITML) (Davis et al., 2007) with feature selection using the maximum-relevance, minimum redundancy criterion (MRMR) (Peng et al., 2005). For the NCA, LFDA, ITML, and MRMR feature selection, we use code provided by the authors. We evaluate the performance of the dimensionality reduction methods via k-NN classification accuracy with $k = 3$, as was done in (Weinberger & Saul, 2009).

As a preprocessing step, we standard normalize the training data so that each feature has a mean of zero and standard deviation of one. We choose the number of neighbors used to estimate the local Gaussians for the LDG dimensionality reduction by five-fold cross-validation using a local QDA classifier (Garcia et al., 2010) on the original data. For LDG, $\gamma \in \{.2, .4, .6, .8, 1\}$, and we choose whichever $\gamma$ min-imizes the k-NN leave-one-out cross-validation error at dimensionality equal to the number of classes plus five. We have found that, in general, a few more dimensions than the number of classes present in the data is a good dimensionality at which to choose $\gamma$. In the case of ties, we select the largest value of $\gamma$. MRMR requires that we discretize the ITML features for feature selection, and we do so by thresholding at the mean, as recommended in the authors' code.

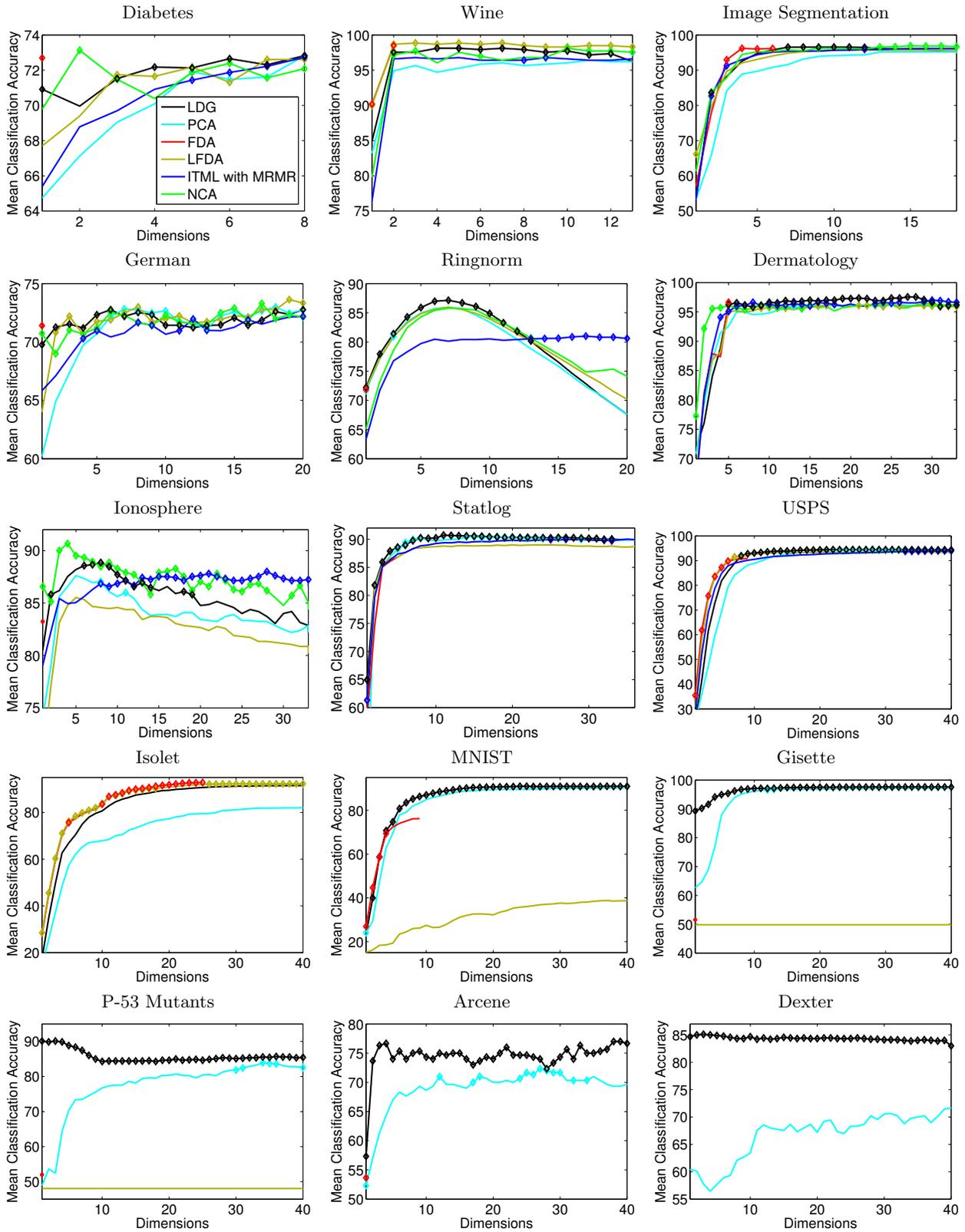

Figure 1. Mean classification accuracy over ten random splits of the data. Diamonds mark methods that are statistically the best or not statistically different from the best with 95% confidence for that dimensionality. FDA, LFDA, ITML, and NCA are not plotted in some datasets for reasons given in Section 4.

Table 1. Mean training time in seconds and mean classification accuracy when the number of dimensions is chosen by cross-validation. Bold font highlights methods that were statistically the best or not statistically different from the best with 95% confidence. Symbol legend: "-" method did not converge in under three hours per training/test split, "nc" not computable.

|  | Orig Dim | Train Ex | LDG acc | LDG time | PCA acc | PCA time | FDA acc | FDA time | LFDA acc | LFDA time | ITML acc | ITML time | NCA acc | NCA time |
|---|---|---|---|---|---|---|---|---|---|---|---|---|---|---|
| Diabetes | 8 | 538 | **71.3** | 1 | 67.9 | <1 | **72.7** | <1 | 69.2 | <1 | 69.7 | 40 | **70.7** | 135 |
| Wine | 13 | 125 | **97.7** | <1 | 95.8 | <1 | **98.5** | <1 | **98.5** | <1 | 97.2 | 54 | **97.9** | 6 |
| Image Seg | 19 | 1617 | **96.5** | 4 | 92.5 | 2 | **96.2** | 1 | 95.0 | 3 | 95.6 | 138 | 95.4 | 4641 |
| German | 20 | 700 | **71.1** | <1 | 68.9 | <1 | **71.4** | <1 | **71.9** | <1 | 69.5 | 15 | **71.1** | 1015 |
| Ringnorm | 20 | 3000 | **86.9** | 9 | 85.8 | 5 | 71.9 | 1 | 85.8 | 6 | 80.6 | 23 | 85.7 | 9119 |
| Derm | 33 | 256 | 89.2 | <1 | 92.5 | <1 | 91.5 | <1 | 92.5 | <1 | 93.4 | 139 | **95.5** | 381 |
| Ion | 34 | 246 | 86.2 | <1 | 85.2 | <1 | 83.2 | <1 | 83.1 | <1 | 84.3 | 10 | **89.1** | 222 |
| Statlog | 36 | 3000 | **90.1** | 10 | **90.1** | 5 | 86.6 | 3 | 88.3 | 6 | 88.0 | 232 | - | - |
| USPS | 256 | 3000 | **93.5** | 24 | 92.0 | 10 | 90.9 | 5 | 92.6 | 7 | 90.8 | 4886 | - | - |
| Isolet | 617 | 3000 | 87.9 | 94 | 73.1 | 15 | **88.9** | 18 | **90.2** | 21 | - | - | - | - |
| MNIST | 784 | 3000 | **89.1** | 55 | 87.5 | 13 | 76.2 | 10 | 32.7 | 34 | - | - | - | - |
| Gisette | 5000 | 3000 | **95.5** | 466 | **96.7** | 63 | 51.5 | 1983 | 49.8 | 2313 | - | - | - | - |
| Mutants | 5408 | 200 | **90.0** | 2908 | 64.8 | 2 | 52.0 | 1946 | 48.0 | 2003 | - | - | - | - |
| Arcene | 10K | 70 | **76.0** | 578 | 61.3 | 15 | 53.7 | 39 | nc | nc | - | - | - | - |
| Dexter | 20K | 210 | **84.0** | 4365 | 58.1 | 2 | nc | nc | nc | nc | - | - | - | - |

We perform experiments on fifteen datasets, and for each we average the accuracy over ten random 70/30 splits of the training and test data (up to a maximum of 3000 training samples). The datasets that we use can be found either at the UCI Machine Learning Repository or the Machine Learning Dataset Repository. The P53-Mutants dataset contained a large degree of class asymmetry. Therefore, we randomly sampled 143 of the *inactive* class samples and discarded the rest in order to make a 50/50 split between *inactive* and *active* class data (as opposed to the 1% vs 99% split in the original dataset).

Figure 1 and Table 1 show that for small datasets, LDG is comparable to other state-of-the-art methods. However, LDG provides a clear advantage on the datasets with the largest feature dimensionality.

NCA and ITML failed to converge in under three hours per training/test split on a standard 2.8 GHz PC for the datasets marked with "-" in Table 1, and results for these datasets are not plotted in Figure 1. Figure 1 also shows that ITML has difficulty with the Ringnorm dataset which has some features that are only noise.

LDG also outperforms FDA and LFDA on some of the datasets. FDA can provide dimensionality only up to one fewer than the number of classes, which limits its performance on the Ionoshpere and Ringnorm datasets. Furthermore, FDA and LFDA exhibit numerical instability in some of the datasets with large feature dimensionality due to the fact that the within-class covariance matrix is underdetermined. Thus, the generalized eigenvalue decomposition that these algorithms solve fails to find discriminative dimensions. LFDA returns complex eigenvalues for the Arcene and Dexter datasets, and FDA does the same on the Dexter dataset; thus, the LFDA and FDA results are not computable for these datasets.

In Table 1, we show the average classification accuracy when the dimensionality is chosen by leave-one-out cross-validation. We do this by increasing the dimensionality until the cross-validation accuracy decreases by adding another dimension. The run-time numbers measure the mean time it takes, in seconds, for the method to produce the dimensions shown in Figure 1 and to select the best dimensionality.

## 5. LDG for Transfer Learning

In this section, we apply LDG dimensionality reduction to transfer learning. In transfer learning, we wish to classify test data drawn from some unknown *target domain* distribution of feature vectors and class labels where we have very few training examples. However, we assume that we have plenty of training examples from a *source domain* that differs from the target domain, but is thought to be useful for learning. For example, in our experiments we treat resized MNIST handwritten digits as the source and USPS handwritten digits as the target (see Figure 4).

Let $\mathcal{T} = \{(x_i, y_i)\}_{i=1}^{n_t}$ be the target domain training data drawn iid from some unknown joint distribution $p_\mathcal{T}(x,y)$. Let $\mathcal{S} = \{(x_\ell, y_\ell)\}_{\ell=n_t+1}^{n_s+n_t}$ be the source domain training data drawn iid from unknown joint distribution $p_\mathcal{S}(x,y) \neq p_\mathcal{T}(x,y)$, with $n_s >> n_t$.

The goal in transfer learning is to achieve high classification accuracy in the target domain by training a classifier using both sets of training data. Therefore, one of the goals for dimensionality reduction is to find a $B$ matrix such that the target domain data are separated according to class. However, we now have the added goal that we wish to find a mapping where the source and target domain distributions are similar, i.e. $p_\mathcal{T}(B^T x, y) \approx p_\mathcal{S}(B^T x, y)$. In Figure 2 we show examples of two different one-dimensional spaces that have been mapped from some higher-dimensional space (which is not shown). The left plot is a mapping in which the target domain data are separated according to class, but the source and target domain distributions are not similar. In the right plot, the target domain data are separated, and additionally, the source domain data distribution is similar to that of the target domain data. Therefore, both the source and target domain data can be used to train a classifier for the test data using the right-side mapping.

For transfer learning, we weight objective (2) for the target and source domain training data using parameter $\alpha$,

$$f(B) = (1-\alpha) \sum_{i=1}^{n_t} \log \left( \frac{\sum_{j=1}^m p(B^T x_i | j) p(j)}{p(B^T x_i | y_i) p(y_i)} \right) \quad (7)$$
$$+ \alpha \sum_{\ell=n_t+1}^{n_s+n_t} \log \left( \frac{\sum_{j=1}^m p(B^T x_\ell | j) p(j)}{p(B^T x_\ell | y_\ell) p(y_\ell)} \right).$$

We estimate the parameters of the Gaussian distribution for *target domain* point $x_i$ using the $k$ nearest *source domain* training examples. The first term in (7) is the primary transfer term. The denominator in this term finds a $B$ that maximizes the likelihood of the target domain data for a Gaussian distribution trained using the local source domain data, thus finding a $B$ that brings the same-class source and target domain data close together. Conversely, the numer-ator in the first term seeks a $B$ that minimizes the likelihood of the different-class source Gaussians, thus ensuring that the mapping is still discriminative.

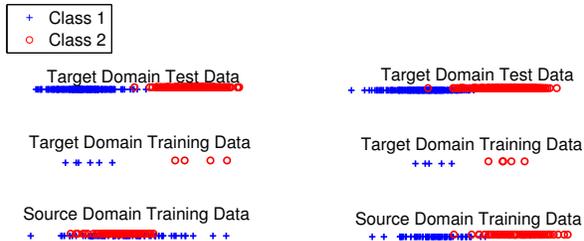

Figure 2. Two examples of one-dimensional mappings for transfer dimensionality reduction. In each example, the target domain is separated, but in the right example the target domain data matches the source domain data, making transfer learning more effective.

The second term in (7) is the normal LDG objective function for the source domain data only. Thus, if $\alpha = 0.5$, (7) is very similar to standard LDG dimensionality reduction acting on the pooled source and target domain data. We include this term because if the source and target domain distributions are similar, then we can set $\alpha = 0.5$ to train $B$ using as much data as possible. We choose $\alpha$ by cross-validating over the target domain training data. In case of ties, we choose the largest value of $\alpha$, thereby defaulting to using as much data as possible. We make the same approximations as in Section 2 to find an analogous approximation to (6) for (7).

## 6. Related Methods for Transfer Learning

Of the dimensionality reduction techniques described in Section 3, only ITML has been adapted to the transfer learning scenario (Saenko et al., 2010). Let $d_M(x_i, x_\ell) = (x_i - x_\ell)^T M (x_i - x_\ell)$, the squared Mahalanobis distance. The original ITML objective is:

$$M^* = \underset{M \succeq 0}{\arg\min} \; \text{Tr}(M) - \log \det(M)$$
$$\text{s.t.} \; d_M(x_i, x_\ell) \leq u, \text{ if } y_i = y_\ell \quad (8)$$
$$d_M(x_i, x_\ell) \geq v, \text{ if } y_i \neq y_\ell.$$

For transfer metric learning, the authors propose to use objective function (8), but generate constraints only between examples from different domains, i.e. $x_i \in \mathcal{T} \; \forall i$ and $x_\ell \in \mathcal{S} \; \forall \ell$. In this way, they find an $M$ that makes distances between examples *across* the two domains small for same-class data and large for different-class data. We again use MRMR feature selection to find a dimensionality reduction matrix from $M$ as described in Section 3.

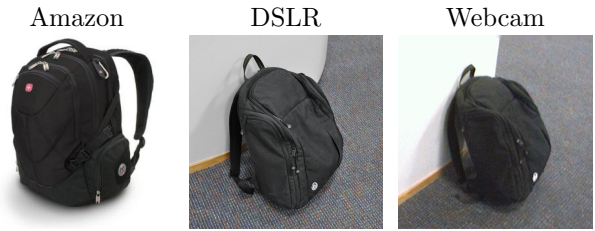

Figure 3. Examples of images taken from the Amazon, DSLR, and Webcam domains.

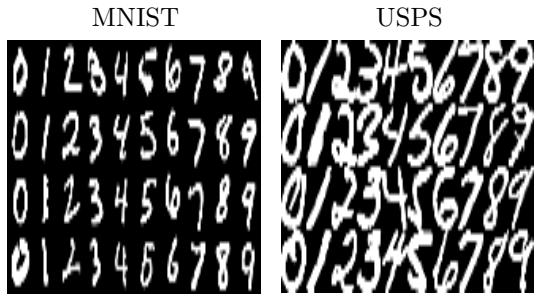

Figure 4. Examples of resized MNIST images and USPS images.

## 7. Transfer Learning Experiments

We conduct transfer learning experiments for two different classification problems. The first is to classify images according to the category of the object found in the images, a thirty class problem, with datasets from three domains: Amazon product images, images taken with a high-resolution DSLR camera, and images taken with a low-resolution webcam. Examples of the back packs category for these three domains is shown in Figure 3. This dataset was first used by Saenko, et al., and we use the same preprocessing techniques as described in (Saenko et al., 2010) to featurize the images, which results in 800 features per image.

In the second problem, the two different domains consist of the grayscale digit images in the MNIST and USPS datasets. The image features are the raw pixel values, and the only preprocessing we use is to resize the MNIST images to 16 x 16 pixels to match the USPS images. We show examples of images from each domain in Figure 4.

We compare four dimensionality reduction techniques by their performance using $k = 3$ nearest-neighbor classification. The first is transfer LDG where we choose $\alpha$ from $[0, .1, .3, .5]$ by k-NN cross-validation at dimensionality equal to the number of classes plus five.

We compare to pooled PCA and pooled FDA dimensionality reduction. These approaches ignore the difference between the domains by pooling the training data from each domain and then performing standard PCA or FDA. Finally, we compare to linear ITML for transfer learning as described in (Saenko et al., 2010) with MRMR feature selection. We also show results for ITML with no dimensionality reduction.

The source domain training data consists of all the available data in that particular domain, and we standard normalize so that it has mean zero and standard deviation one. The target domain training data consists of exactly two examples per class and is standard normalized independently of the source domain training data. We remove any features that exhibit zero variance in the source or target domain.

Figure 5 plots the accuracy averaged over ten random splits of the target domain test and training data. The results show that LDG is statistically the best or tied for the best at many dimensions in all experiments. Pooled PCA performs well in the datasets where Amazon images act as the source, but fails to perform as well as LDG on the other datasets. Pooled FDA performs poorly on all of the datasets.

We do not show results for the best dimensionality chosen by cross-validation, similar to those in Table 1, due to space constraints. We do note that for the dimensions we have plotted, ITML achieves its highest accuracy with no dimensionality reduction, but still does not match the best performance of LDG.

## 8. Conclusions

We have presented LDG dimensionality reduction, a technique that maps the data to a space where classes are separated locally to each training point. LDG is solved via a simple maximal eigenvalue decomposition, and thus scales better than iterative methods and LFDA for large datasets. Furthermore, we have shown that LDG dimensionality reduction can be applied to transfer learning problems with good results.

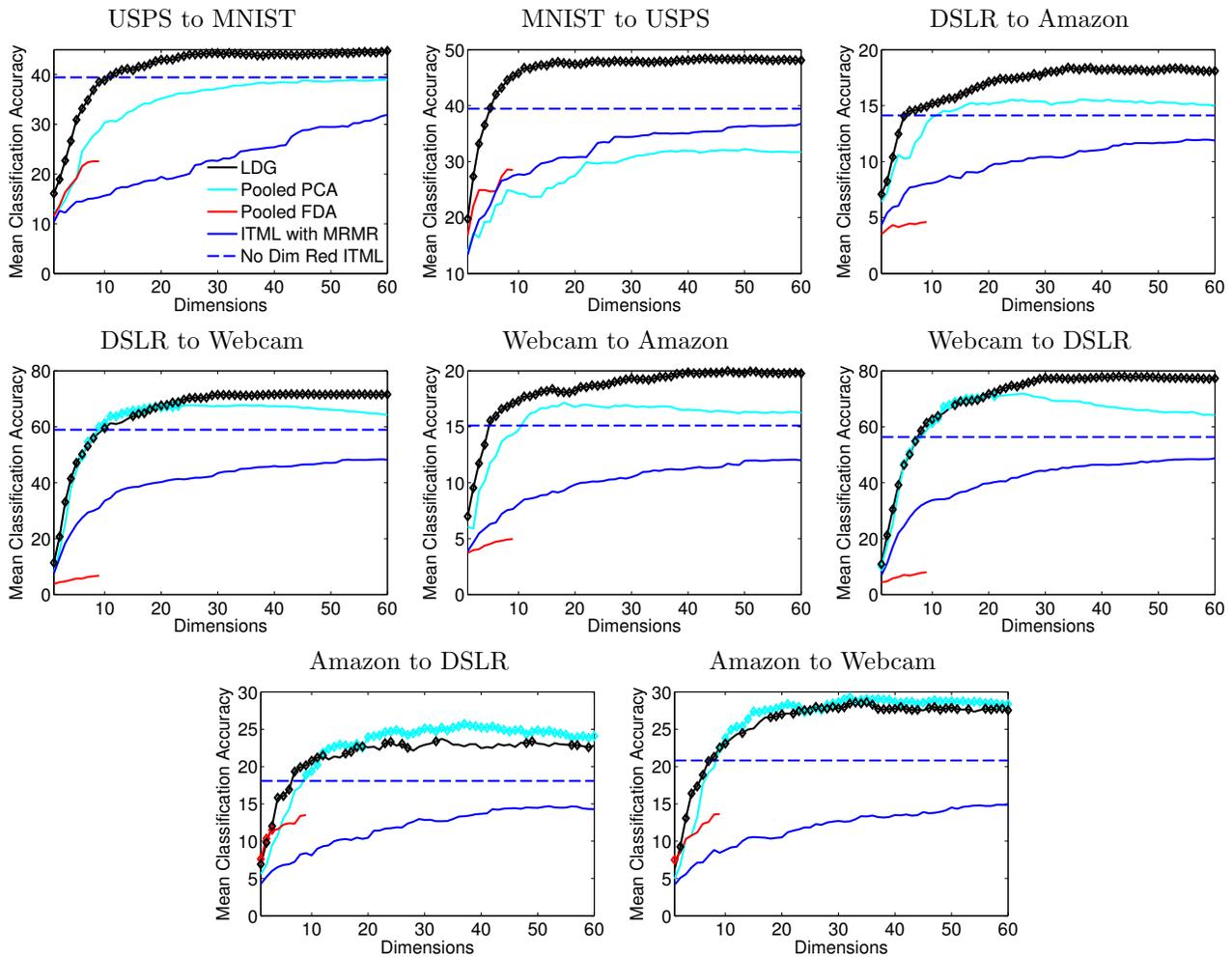

*Figure 5.* Transfer dimensionality reduction results when we randomly sample exactly two target domain training examples per class. Diamonds indicate that the method was statistically the best or not statistically different from the best with 95% confidence for that dimensionality.